\documentclass[runningheads]{llncs}

\usepackage[mobile]{eccv}

\usepackage{eccvabbrv}

\usepackage{graphicx}
\usepackage{booktabs}

\usepackage[accsupp]{axessibility}  % Improves PDF readability for those with disabilities.

\usepackage[pagebackref,breaklinks,colorlinks,citecolor=eccvblue]{hyperref}
\usepackage{hyperref}
\usepackage{multicol}
\usepackage{multirow}
\usepackage{orcidlink}
\usepackage{graphicx}  % scalebox
\usepackage{multirow}  % multirow
\usepackage{xcolor}    % color
\usepackage{bm}        % bm (bold math)
\usepackage[table]{xcolor}
\usepackage{subcaption}

\usepackage{pifont}
\newcommand{\cmark}{\ding{51}} % checkmark
\newcommand{\xmark}{\ding{55}} % cross

\begin{document}

\title{Odoriko: A Shape-Aware Multimodal Diffusion Framework for Human Motion} 

\titlerunning{Odoriko}

% Second Author\inst{2,3}\orcidlink{1111-2222-3333-4444} \and
% Third Author\inst{3}\orcidlink{2222--3333-4444-5555}}

\author{
Dongseok Shim\inst{1}\and Julian Tanke\inst{2} \and
Kengo Uchida\inst{2} \and Christian Simon\inst{1} \and 
Koichi Saito\inst{2} \and Takashi Shibuya\inst{2} \and
Shusuke Takahashi\inst{1} \and Yuki Mitsufuji \inst{1, 2}
}

\authorrunning{D. Shim \etal}

\institute{ $^{1}$Sony Group Corporation \quad $^{2}$Sony AI
\\ \email{first\_name.last\_name@sony.com} }

\maketitle

\begin{abstract}

Human motion generation has been widely studied across diverse input modalities — text, music, and video — and recent efforts have unified these into single multimodal frameworks. However, while morphological factors such as gender and body shape are known to produce distinct kinematic signatures, no existing unified framework incorporates this into generation, treating all subjects as morphologically equivalent.
We present Odoriko, the first unified multimodal motion generation framework that reflects subject bio-morphological information directly in synthesized motion output. Rather than averaging over subject variation, Odoriko generates motion that is consistent with who is moving, not just what they are asked to do — across text, music, and video conditions within a single model. 
When explicit morphological information is unavailable, Odoriko additionally recovers subject morphology alongside motion, unifying estimation and generation in one framework.
Extensive experiments across text-to-motion, music-to-dance, and video-to-motion benchmarks demonstrate that Odoriko matches or exceeds prior specialized models on standard metrics, while enabling morphology-consistent generation that no existing unified framework supports.
Project page: \url{https://dsshim0125.github.io/odoriko.github.io/}

\end{abstract}

\section{Introduction}

Human motion synthesis and reconstruction are central to animation, VR, robotics, gaming, and embodied AI~\cite{holden2017phase, parent2012computer, starke2021neural, yamane2013synthesizing, li2023object}.
Recent advances in generative modeling have enabled realistic motion generation from diverse inputs such as text, music, video, and sparse keypoints, lowering the barrier for content creation and enabling richer human–computer interaction.
\begin{figure}
    \centering
    \includegraphics[width=0.8\linewidth]{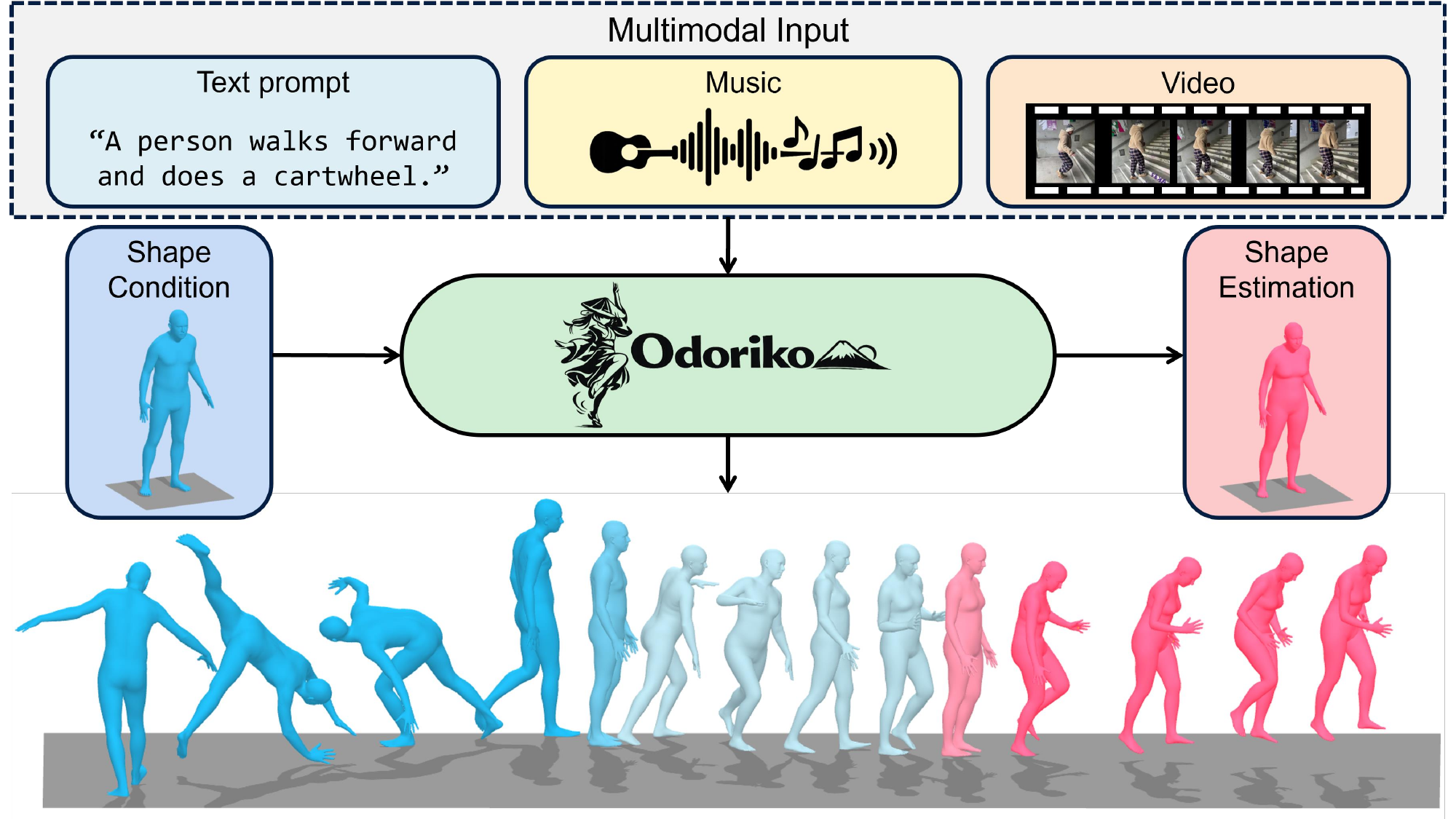}
    \caption{
    We present Odoriko, a shape-aware multimodal human motion framework that unifies motion generation and 3D pose estimation within a single model.
    Our approach explicitly models subject-specific body shape, enabling both shape-conditioned motion synthesis and shape-aware pose estimation under diverse multimodal inputs.
    }
    \label{fig:fig1}
    \vspace{-20pt}
\end{figure}

Despite this rapid progress, current motion generation systems remain fundamentally limited in two critical aspects.
First, most models are designed as single-modality specialists, excelling at one specific task—such as text-to-motion \cite{tevet2022human, zhang2023generating, dai2024motionlcm, pinyoanuntapong2024mmm, guo2024momask, uchida2025mola}, music-to-dance \cite{li2021ai, kim2022brand, siyao2022bailando, li2024lodge, li2024lodge++}, or video-based pose reconstruction \cite{sun2023trace, shin2024wham, shen2024world, wang2024tram}—while failing to generalize across modalities.
This fragmented design requires separate architectures and training pipelines for each modality, preventing unified control across tasks.
Second, and more critically, existing approaches largely overlook the subject-specific nature of human motion.
Human motion is not only a function of action intent or external conditions, but of bio-morphological characteristics of the subject such as limb proportions, mass distribution, and gender-specific structure—which systematically influence kinematics and dynamics\cite{barclay1978temporal, cutting1978generation, troje2002decomposing, hsue2014effects}.
For example, the same textual instruction or musical rhythm should yield systematically different motions when performed by subjects with different body shapes.

Recent works have begun incorporating body shape into motion synthesis~\cite{arbol2024bodyshapegpt, choutas2022accurate, liao2025shape, wang2025generating}, highlighting the importance of morphology-aware modeling. 
However, most approaches focus on a single modality and rely on handcrafted or static shape representations, which limit their ability to capture rich subject-specific variation. 
Moreover, as they are typically designed around one input modality, they do not naturally extend to heterogeneous conditions such as music or video.

Yet, extending shape-aware modeling across modalities is inherently non-trivial, since the influence of body morphology on motion depends on context—for instance, affecting rhythmic dynamics in music-driven dance differently from semantic articulation in text-based actions. 
Modeling these modality-dependent shape–motion interactions within a unified framework therefore remains relatively underexplored, motivating the need for a general multimodal, shape-aware motion system.

In this work, we present \textbf{Odoriko}, a unified framework for shape-aware multimodal human motion generation and reconstruction. To the best of our knowledge, Odoriko is the first model to explicitly incorporate subject-specific shape factors either through conditioning or estimation across multiple input modalities, and to consistently reflect them in the generated motion. 
The framework supports text-to-motion synthesis, music-to-dance generation, and local human motion estimation from videos and 2D keypoints within a single shared architecture, enabling controllable and anatomically coherent motion modeling across tasks and modalities.

We represent subject shape using gender and SMPL \cite{loper2015smpl} $\boldsymbol{\beta}$ parameters, providing a compact yet expressive encoding of body morphology. 
To integrate shape with multimodal conditions, we propose \emph{Shape-Aware Motion Transformer (SAMT)} within a diffusion-based framework, injecting shape cues at multiple network levels to modulate motion dynamics according to subject morphology. Odoriko further leverages pretrained modality-specific encoders, decoupling feature extraction from motion synthesis and enabling the generator to focus on learning shape-conditioned motion distributions. 

We evaluate Odoriko on standard text-to-motion, music-to-dance, and video-to-motion (local human pose estimation) benchmarks, and introduce new protocols to quantify shape fidelity in generated motions. 
Results show that Odoriko matches or surpasses state-of-the-art performance in motion quality and diversity while more faithfully capturing subject-specific shape characteristics.

\begin{figure}[t]
    \centering
    \includegraphics[width=1.0\linewidth]{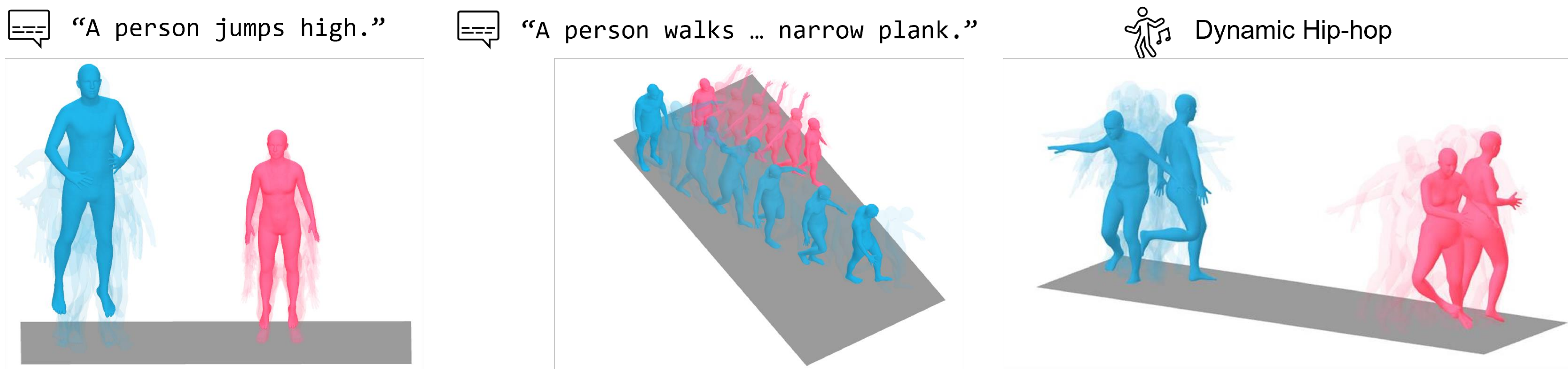}
    \caption{Odoriko visualizations showing shape-dependent motion variations under identical text or music conditions.}
    \vspace{-20pt}
    \label{fig:qual}
\end{figure}

In short, our contribution can be summarized as follows.
\begin{itemize}
    \item We introduce \textbf{Odoriko}, the first unified framework for shape-aware multimodal human motion generation and reconstruction, capable of handling diverse input modalities within a single model.
    \item We propose the Shape-Aware Motion Transformer (SAMT) that effectively injects subject shape cues jointly with multimodal conditioning signals.
    \item Extensive experiments across multiple benchmarks, including text-to-motion, music-to-dance, and local human motion estimation, demonstrate that Odoriko achieves competitive or superior performance while generalizing across input modalities.
    \item We introduce shape-aware motion generation benchmarks that evaluate subject–motion consistency, demonstrating Odoriko’s superiority in preserving shape-dependent motion patterns and highlighting the importance of shape awareness in human motion generation and estimation.
\end{itemize}

\section{Related Work}

\noindent \textbf{Human Motion Generation and Estimation.}
Text-driven motion synthesis has advanced rapidly through deep generative models.
Diffusion-based methods~\cite{tevet2022human, zhang2024motiondiffuse} progressively denoise motion representations to produce semantically aligned sequences, while latent variants~\cite{chen2023executing, dai2024motionlcm} improve efficiency by operating in compact latent spaces.
Discrete approaches leveraging VQ-VAE pretraining~\cite{pinyoanuntapong2024mmm, guo2024momask} and masked autoregressive modeling~\cite{meng2025rethinking} further improve quality and long-range temporal coherence.
For music-conditioned dance generation, methods have evolved from autoregressive Transformers~\cite{li2021ai} and GAN-based models~\cite{siyao2022bailando} to diffusion-based frameworks with strong musical alignment~\cite{tseng2023edge, li2023finedance, li2024lodge}.
In contrast, video-based motion estimation recovers 3D pose trajectories from observations rather than synthesizing them; recent feed-forward frameworks~\cite{shin2024wham, shen2024world, wang2024tram} achieve high accuracy without generative pipelines.
Unlike these single-task methods, our work unifies all three paradigms, text-to-motion, music-to-dance, and video-to-motion estimation, within a single model.

\noindent \textbf{Multimodal Motion Synthesis.}
Building generalist motion models over heterogeneous input modalities is an emerging direction.
M$^3$GPT~\cite{luo2024m} integrates motion and language via instruction tuning on large language models.
MCM~\cite{ling2024mcm} and MotionCraft~\cite{bian2025motioncraft} address optimization conflicts in mixed-modality training via ControlNet-based conditioning and coarse-to-fine training strategies, respectively.
Most closely related to our work, GENMO~\cite{li2025genmo} unifies motion estimation and generation under a single diffusion framework, supporting diverse conditioning signals including text, music, and video.
MotionLab~\cite{guo2025motionlab} further unifies generation and editing tasks under a Motion-Condition-Motion paradigm using rectified flows.
However, none of these methods incorporate subject-specific body shape as a conditioning signal, nor do they enforce biomechanically consistent motion across diverse body morphologies—both of which are central to our approach.

\noindent \textbf{Shape-Aware Motion Modeling.}
Biological morphology—mass distribution, limb proportions, and joint torque limits—fundamentally determines how individuals move~\cite{barclay1978temporal, cutting1978generation, troje2002decomposing, hsue2014effects}.
Ignoring shape leads to physically inconsistent outputs such as floating, ground penetration, or biomechanically implausible postures.
Several works integrate shape parameters into pose estimation and reconstruction with physically grounded constraints~\cite{choutas2022accurate, gralnik2023semantify, arbol2024bodyshapegpt}, and HUMOS~\cite{tripathi2024humos} introduces a shape-conditioned generative model using cycle consistency and differentiable physics losses.
More recently, ShapeMove~\cite{liao2025shape} proposes a text-driven shape-aware generation framework combining a dedicated FSQ-VAE with a pretrained language model, and AttrMoGen~\cite{wang2025generating} introduces attribute-controlled motion generation by disentangling action semantics from body attributes via a structural causal model.
Despite these advances, shape-aware motion modeling in \emph{multimodal} generation settings---where driving signals such as music or text carry no identity information---remains largely unaddressed.
Our work is the first to tackle this challenge in a unified multi-task framework spanning text, audio, and video conditioning.
\begin{figure}[t]
    \centering
    \includegraphics[width=1.0\linewidth]{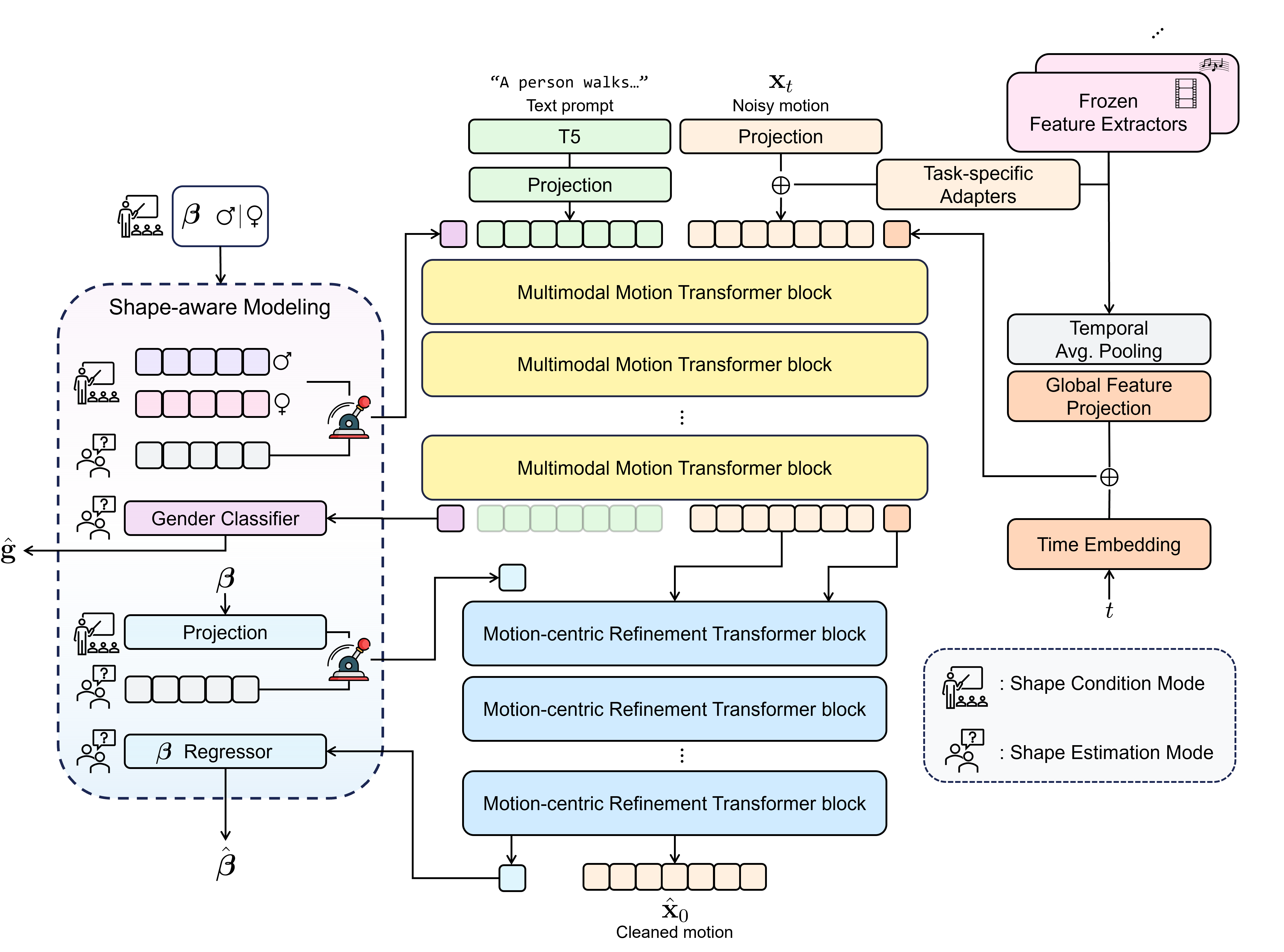}
\caption{
\textbf{Odoriko Overview.}
Temporally aligned multimodal inputs are fused with noisy motion tokens, while their global features condition the denoising timestep $t$.
A shape token constructed from gender $\mathbf{g}$ and body shape $\boldsymbol{\beta}$ is integrated via joint attention, enabling explicit morphology-aware generation. 
The model outputs denoised motion $\hat{\mathbf{x}}_{0}$, and additionally predicts shape parameters in shape estimation mode. 
For clarity, the CLIP text embedding path is omitted.
}
\vspace{-20pt}
\label{fig:odoriko}
\end{figure}

\vspace{-10pt}
\section{Method}

In this section, we present \textbf{Odoriko}, a unified and shape-aware framework for multimodal human motion modeling. 
Our approach enables the generation of diverse and realistic motion sequences conditioned on heterogeneous inputs, including text, music, video, and 2D keypoints. 

Beyond multimodal controllability, our framework explicitly models subject-specific characteristics by incorporating intrinsic biomechanical and kinematic properties of the motion subject. 
This design allows the generated motions to faithfully reflect variations in body shape, gender, and physical structure, leading to improved physical plausibility and subject consistency across different generation scenarios.

\subsection{Motion Representation}
\label{sec31}

Human motion can be represented in various forms, including parametric body models such as SMPL~\cite{loper2015smpl}, dataset-specific formats such as HumanML3D~\cite{guo2022generating}, and discrete or latent motion codes adopted in recent generative frameworks~\cite{guo2024momask, meng2025rethinking}. The choice of representation plays a critical role in model stability, physical plausibility, and learning efficiency, particularly for diffusion-based generative models.

We directly utilize the articulated pose parameters and transform the global transformation into a rotation and translation-invariant canonical frame~\cite{holden2016deep,petrovich2024multi}.
Each motion sequence $\mathbf{x}$ is represented as:
\begin{equation}
\mathbf{x} = \{ r_z,\ \dot{r}_x,\ \dot{r}_y,\ \dot{\alpha},\ \boldsymbol{\theta} \},
\end{equation}
where $r_z$ denotes the root (pelvis) height from the ground, assuming the $xy$-plane corresponds to the ground plane and the $z$-axis points upward. 
The terms $\dot{r}_x$ and $\dot{r}_y$ represent the root linear velocities in the ground plane, and $\dot{\alpha}$ denotes the angular velocity around the vertical ($z$) axis. 
The variable $\boldsymbol{\theta}$ corresponds to local joint rotations derived from SMPL body pose parameters. To ensure continuity and avoid singularities inherent in Euler angles or quaternions, joint rotations are encoded using a continuous 6D rotation representation~\cite{zhou2019continuity}.

To explicitly model camera-centric orientation when required (e.g., in human pose estimation settings), we introduce an auxiliary parameter $\boldsymbol{\theta}_{r}^{l} \in \mathbb{R}^{6}$ that represents the camera-aligned root orientation in a 6D rotation formulation. 
This auxiliary parameter is used exclusively for camera-dependent tasks such as pose estimation, and is omitted in generation scenarios (e.g., text-to-motion) where camera alignment is not required.

\subsection{Data-Domain Diffusion Framework}

We adopt a data-domain diffusion formulation \cite{ho2020denoising} with direct $\mathbf{x}_0$ prediction, following prior human motion synthesis works \cite{tevet2022human, zhang2023remodiffuse, ren2024realistic, li2025genmo, li2024lodge, li2024lodge++}.
Human motion lies on a highly structured, physically constrained manifold characterized by strong inter-joint correlations and temporal coherence. Modeling diffusion directly in the pose space preserves these geometric and kinematic dependencies, facilitating anatomically consistent and dynamically plausible motion reconstruction.

Accordingly, we employ the standard $\mathbf{x}_0$-estimation objective in the data domain:
\begin{equation}
\hat{\mathbf{x}}_{0} = f_{\text{Odoriko}}(\mathbf{x}_{t}, t, \mathbf{c}, \mathbf{s}),
\end{equation}
where $\mathbf{x}_{t}$ denotes the noisy motion representation at timestep $t$, $\mathbf{c}$ represents multimodal conditioning signals (e.g., text, music, video, or 2D keypoints), and $\mathbf{s}$ encodes subject-specific shape factors, further described in Sec. \ref{subsec:arc}.

\subsection{Model Architecture}
\label{subsec:arc}
In this section, we present the architecture of \textbf{Odoriko}, illustrated in Fig.~\ref{fig:odoriko}. 
Odoriko is built upon the proposed \emph{Shape-Aware Motion Transformer} (SAMT), a diffusion-based architecture that explicitly incorporates subject-specific body shape information to enable morphology-aware motion modeling from multimodal inputs.  
The detailed design of SAMT is described below.

\vspace{4pt}
\noindent
\textbf{Multimodal Conditioning.}
Odoriko supports heterogeneous inputs, including text, music, video, and 2D keypoints. 
To decouple representation learning from motion synthesis and ensure stable optimization, we adopt modality-specific \emph{frozen} pre-trained feature extractors for all modalities except 2D keypoints. 
This design allows the diffusion backbone to focus on motion dynamics while leveraging semantically rich representations from large-scale foundation models.

For text encoding, we employ T5-Base~\cite{raffel2020exploring} to extract token-level contextual embeddings that capture fine-grained linguistic structure, and CLIP~\cite{radford2021learning} to obtain sentence-level global representations that encode holistic semantic meaning.

For music, audio waveforms are processed by the Jukebox encoder~\cite{dhariwal2020jukebox}, followed by the EDGE transformer~\cite{tseng2023edge}, capturing rhythmic structure and high-level musical semantics. 
Music inputs are temporally resampled to align with the target motion sequence.

For video and 2D keypoints, video features are extracted using the frozen TRAM encoder~\cite{wang2024tram}, producing spatially compressed linearized visual embeddings. 
For 2D keypoints, we extract 18-joint OpenPose-style detections~\cite{cao2019openpose} using DWPose~\cite{yang2023effective} and project them into the motion embedding space via lightweight MLP adapters.

Except for text, all modality features are temporally aligned to the motion sequence through resampling or linear interpolation. 

\vspace{4pt}
\noindent
\textbf{Multimodal Motion Transformer.}
The diffusion backbone of \textbf{Odoriko} builds upon the Multimodal Diffusion Transformer (MM-DiT)~\cite{esser2024scaling}, which provides a unified self-attention framework for modeling heterogeneous conditioning signals. 

Following MM-DiT, text tokens from T5 are concatenated with motion tokens along the temporal axis, forming a unified token sequence processed via full self-attention. 
For temporally aligned modalities (music, video, and 2D keypoints), modality-specific features are projected into the motion embedding space through lightweight MLP adapters and added to the corresponding motion tokens. This additive fusion preserves temporal synchronization while maintaining parameter efficiency and stable diffusion training. 
Also, to handle variable-length motion sequences, we zero-pad shorter sequences and apply masked self-attention to ignore padded positions. For video-to-motion tasks that do not require textual input, text tokens are similarly masked out, so attention is restricted to valid modality tokens only.

\vspace{4pt}
\noindent
\textbf{Global Sequence-Level Conditioning.}
Beyond token-level cues, we incorporate a sequence-level global signal to promote long-range coherence.
For all modalities except text, a global feature is obtained by temporally pooling frame-wise features from the respective frozen encoder, followed by a linear projection; for text, we directly use the CLIP sentence-level embedding, which already provides a holistic global representation.
The resulting vector is added to the denoising timestep embedding and prepended to the motion token sequence as a global conditioning token, anchoring semantic consistency across the full denoising trajectory.

\vspace{4pt}
\noindent
\textbf{Efficient Hybrid Transformer Design.}
To balance multimodal modeling capacity and computational efficiency, we adopt a staged hybrid transformer architecture inspired by FLUX \cite{labs2025flux1kontextflowmatching} and MMAudio \cite{cheng2025mmaudio}. 
The network is divided into two functional halves with distinct roles.

The first half consists of Multimodal Motion Transformer blocks, where motion tokens attend jointly to text tokens, global conditioning tokens, and a subject-level shape token introduced later. These layers establish cross-modal semantic grounding while preserving motion structure.

The second half comprises Motion-Centric Refinement Transformer blocks. After semantic alignment is established, text tokens are removed to reduce attention complexity, and motion tokens are further refined using contextual information from global conditioning tokens and the shape token.

\vspace{4pt}
\noindent
\textbf{Shape-Aware Modeling.}
A central contribution of Odoriko is the explicit integration of human body shape into multimodal motion modeling. 
We represent subject morphology using SMPL parameters
\(
\mathbf{s} = \{\mathbf{g}, \boldsymbol{\beta}\},
\)
where biological gender \(\mathbf{g} \in \{\text{male}, \text{female}\}\) selects a template and blendshape basis, and body shape coefficients \(\boldsymbol{\beta} \in \mathbb{R}^{10}\) describe continuous deformations within that basis.

Importantly, SMPL exhibits a hierarchical dependency: gender determines the underlying template and basis, while \(\boldsymbol{\beta}\) parameterizes variations within that template. 
We therefore align shape conditioning with this structure through staged injection.

\vspace{2pt}
\noindent
\emph{Multimodal stage (gender conditioning).}
In the  Multimodal Motion blocks, we introduce a gender token derived from a learnable embedding lookup. 
Since early layers perform cross-modal grounding, conditioning them on \(\mathbf{g}\) provides a stable template-level identity prior during semantic alignment.

\vspace{2pt}
\noindent
\emph{Refinement stage (\(\boldsymbol{\beta}\) conditioning).}
In the Motion-Centric Refinement blocks, we inject a \(\boldsymbol{\beta}\) token obtained by projecting the raw $\boldsymbol{\beta}\in\mathbb{R}^{10}$ through an MLP and linear layer. 
Because later layers refine detailed kinematics, conditioning them on \(\boldsymbol{\beta}\) enables modulation of fine-grained biomechanical characteristics.

\vspace{4pt}
\noindent
\textbf{Operational Modes.}
Odoriko operates under a unified formulation with two modes that differ only in how the shape token is instantiated.
In \texttt{shape conditioning}, a subject-level shape token is constructed from provided gender $\mathbf{g}$ and SMPL shape coefficients $\boldsymbol{\beta}$, and injected into the designated transformer stages to enforce morphology-consistent motion synthesis.
In \texttt{shape estimation}, the explicit shape token is replaced by a learnable estimation token; the corresponding transformer output is decoded to predict $\hat{\mathbf{g}}$ and $\hat{\boldsymbol{\beta}}$ via a lightweight classifier and regressor, enabling joint motion reconstruction and implicit morphology inference.

By default, Odoriko operates in \texttt{shape conditioning} for generation (text-to-motion, music-to-dance) and in \texttt{shape estimation} for reconstruction/estimation (video-to-motion).
When ground-truth shape annotations are unavailable (e.g., FineDance~\cite{li2023finedance}), we replace explicit shape inputs with a learnable placeholder token; thus, Odoriko can be applied to such datasets even for generation, in a shape-agnostic manner, without modifying the architecture.

\subsection{Loss Function and Inference}

Following Sec.~\ref{sec31}, Odoriko adopts direct $\mathbf{x}_{0}$ prediction as the diffusion parameterization.
Given a noised sample $\mathbf{x}_{t} \sim q(\mathbf{x}_{t} \mid \mathbf{x}_{0})$ at timestep $t \sim \mathcal{U}(0, T)$, 
the network predicts the clean motion $\hat{\mathbf{x}}_{0}$.
We optimize the standard DDPM objective~\cite{ho2020denoising}:
\begin{equation}
 \mathcal{L}_{\mathrm{diff}} 
 =
 \mathbb{E}_{t,\, \mathbf{x}_{t}}
 \left[
 \left\|
 \mathbf{x}_{0} - \hat{\mathbf{x}}_{0}
 \right\|^{2}
 \right].
\end{equation}
This loss directly supervises reconstruction of the clean motion signal and yields stable training with high sample fidelity.

\vspace{4pt}
\noindent
\textbf{Shape-Aware Auxiliary Supervision.}
When operating in \texttt{shape estimation}, Odoriko additionally predicts biological gender $\hat{\mathbf{g}}$ and SMPL body shape coefficients $\hat{\boldsymbol{\beta}}$ from the global motion representation.
We apply auxiliary supervision:
\begin{equation}
 \mathcal{L}_{\beta}
 =
 \left\|
 \boldsymbol{\beta} - \hat{\boldsymbol{\beta}}
 \right\|^{2},
 \qquad
 \mathcal{L}_{g}
 =
 \mathrm{CE}(\mathbf{g}, \hat{\mathbf{g}}),
\end{equation}
where $\mathrm{CE}(\cdot)$ denotes cross-entropy for gender classification.
The overall objective is
\begin{equation}
 \mathcal{L}_{total}
 =
 \mathcal{L}_{\mathrm{diff}}
 + \lambda_{\beta} \mathcal{L}_{\beta}
 + \lambda_{g} \mathcal{L}_{g}.
\end{equation}
We set the auxiliary weights by mode: in \texttt{shape conditioning}, shape inputs are provided and we disable auxiliary prediction losses ($\lambda_{\beta}=\lambda_{g}=0$); in \texttt{shape estimation}, we enable auxiliary supervision and set $\lambda_{\beta}=\lambda_{g}=0.1$ to balance motion reconstruction and shape supervision.

\vspace{4pt}
\noindent
\textbf{Inference.}
In \texttt{shape conditioning}, user-specified $\mathbf{g}$ and $\boldsymbol{\beta}$ are encoded into the shape token and injected into the diffusion backbone to guide reverse denoising.
In \texttt{shape estimation}, the network predicts $\hat{\mathbf{g}}$ and $\hat{\boldsymbol{\beta}}$ alongside motion reconstruction at each denoising step.
To obtain stable final estimates, we aggregate the timestep-wise predictions by averaging them over the entire denoising trajectory of $T$ steps for robust shape estimation:\[
\hat{\boldsymbol{\beta}} = \frac{1}{T} \sum_{t=1}^{T} \hat{\boldsymbol{\beta}}^{(t)}, 
\qquad
\hat{\mathbf{g}} = \frac{1}{T} \sum_{t=1}^{T} \hat{\mathbf{g}}^{(t)}.
\]
For efficient and accurate sampling, we adopt the UniPC sampler~\cite{zhao2023unipc}, a higher-order predictor--corrector solver, instead of conventional DDPM or DDIM~\cite{songdenoising} sampling schemes.

\section{Experiment}
\subsection{Experiment Setup}
\noindent
\textbf{Datasets.}
\noindent
\emph{Text-to-Motion.}
We use the HumanML3D dataset~\cite{guo2022generating}, constructed from AMASS~\cite{mahmood2019amass} and HumanAct12~\cite{guo2020action2motion}. 
Since our model conditions on body shape parameters ($\boldsymbol{\beta}$), we bypass the retargeting step applied in the original dataset construction and train directly on the AMASS subset, which retains ground-truth SMPL gender and shape annotations; HumanAct12 lacks such annotations and is excluded.

\vspace{2pt}
\noindent
\emph{Music-to-Dance.}
We primarily train on FineDance~\cite{li2023finedance} to leverage long, high-quality dance sequences, and additionally use AIST++~\cite{li2021ai} to provide explicit shape supervision via dancers' gender annotations.

\vspace{2pt}
\noindent
\emph{Video-to-Motion.}
We train on EMDB~\cite{kaufmann2023emdb}, 3DPW~\cite{von2018recovering}, and Human3.6M~\cite{ionescu2013human3}, excluding EMDB Sequence 1 as it is reserved for evaluation. 2D keypoints are extracted with DWPose~\cite{yang2023effective} for both training and inference.

\vspace{4pt}
\noindent
\textbf{Evaluation Metrics.}
Following standard protocols, we report Top-3 R-Precision, FID, MMDist, and Diversity~\cite{guo2022generating, tevet2022human} for text-to-motion; FID, Diversity, and BAS~\cite{li2021ai, li2023finedance, li2024lodge} for music-to-dance; and PA-MPJPE, MPJPE, PVE, and Acceleration error~\cite{sun2023trace, wang2024tram} for video-to-motion.

\begin{table*}[t]
    \centering

\caption{Comparison with state-of-the-art methods on the HumanML3D benchmark. Odoriko is compared primarily against generalist models that handle multiple tasks and modalities. Task-specific text-to-motion methods are shown in {\color{gray}gray} for reference. \colorbox{red!13}{Best} and \colorbox{yellow!16}{second-best} results are highlighted.}
\vspace{-10pt}

    \resizebox{\textwidth}{!}{

\begin{tabular}{clcccccccc}
\toprule
\multirow{2}{*}{Category} 
& \multirow{2}{*}{Method} 
& \multirow{2}{*}{\#param} 
& \multicolumn{3}{c}{R-Precision $\uparrow$} 
& \multirow{2}{*}{FID $\downarrow$} 
& \multirow{2}{*}{MMDist $\downarrow$} 
& \multirow{2}{*}{Diversity $\rightarrow$} 
 \\ 
\cmidrule{4-6}
 & & & Top-1 & Top-2 & Top-3 & & & &  \\
\midrule

N/A 
& Real 
& - 
& $0.511^{\pm.003}$ 
& $0.703^{\pm.003}$ 
& $0.797^{\pm.002}$ 
& $0.002^{\pm\mathrm{N/A}}$ 
& $2.974^{\pm.008}$ 
& $9.503^{\pm.065}$ 
  \\ 
\midrule

\multirow{8}{*}{\textbf{Text-only}} 
& \color{gray}MotionDiffuse~\cite{zhang2024motiondiffuse} 
& \color{gray}{87M}
& \color{gray}$0.491^{\pm.001}$ 
& \color{gray}$0.681^{\pm.001}$ 
& \color{gray}$0.782^{\pm.001}$ 
& \color{gray}$0.630^{\pm.001}$ 
& \color{gray}$3.113^{\pm.001}$ 
& \color{gray}$9.410^{\pm.049}$  \\

& \color{gray}MDM~\cite{tevet2022human}      
& \color{gray}{18M}
& \color{gray}$0.320^{\pm.005}$ 
& \color{gray}$0.498^{\pm.004}$ 
& \color{gray}$0.611^{\pm.007}$ 
& \color{gray}$0.544^{\pm.044}$ 
& \color{gray}$5.566^{\pm.027}$ 
& \color{gray}$9.559^{\pm.086}$ \\

& \color{gray}MLD~\cite{chen2023executing}    
& \color{gray}{26M}
& \color{gray}$0.481^{\pm.003}$ 
& \color{gray}$0.673^{\pm.003}$ 
& \color{gray}$0.772^{\pm.002}$ 
& \color{gray}$0.473^{\pm.013}$ 
& \color{gray}$3.196^{\pm.010}$  
& \color{gray}$9.724^{\pm.082}$    \\

& \color{gray}T2M-GPT~\cite{zhang2023generating} 
& \color{gray}{248M}
& \color{gray}$0.492^{\pm.003}$ 
& \color{gray}$0.679^{\pm.002}$ 
& \color{gray}$0.775^{\pm.002}$ 
& \color{gray}$0.141^{\pm.005}$ 
& \color{gray}$3.121^{\pm.009}$ 
& \color{gray}$9.722^{\pm.081}$   \\

& \color{gray}MotionLCM~\cite{dai2024motionlcm}    
& \color{gray}{40M}
& \color{gray}$0.502^{\pm.003}$ 
& \color{gray}$0.698^{\pm.002}$ 
& \color{gray}$0.798^{\pm.002}$ 
& \color{gray}$0.304^{\pm.012}$ 
& \color{gray}$3.012^{\pm.007}$  
& \color{gray}$9.607^{\pm.066}$    \\

& \color{gray}MMM~\cite{pinyoanuntapong2024mmm}          
& \color{gray}{254M}
& \color{gray}$0.504^{\pm.003}$ 
& \color{gray}$0.696^{\pm.003}$ 
& \color{gray}$0.794^{\pm.002}$ 
& \color{gray}$0.080^{\pm.003}$ 
& \color{gray}$2.998^{\pm.007}$ 
& \color{gray}$9.411^{\pm.058}$ \\

& \color{gray}MoMask~\cite{guo2024momask}       
& \color{gray}{45M}
& \color{gray}$0.521^{\pm.002}$ 
& \color{gray}$0.713^{\pm.002}$ 
& \color{gray}$0.807^{\pm.002}$ 
& \color{gray}$0.045^{\pm.002}$ 
& \color{gray}$2.958^{\pm.008}$ 
& \color{gray}-  \\

& \color{gray}MoLA~\cite{uchida2025mola} 
& \color{gray}{49M}
& \color{gray}$0.516^{\pm.006}$ 
& \color{gray}$0.712^{\pm.005}$ 
& \color{gray}$0.805^{\pm.004}$ 
& \color{gray}$0.115^{\pm.004}$ 
& \color{gray}$3.008^{\pm.016}$ 
& \color{gray}$9.885^{\pm.152}$ \\

\midrule
\multirow{6}{*}{\textbf{Multimodal}}
& TM2D~\cite{gong2023tm2d}
& 72M
& $0.319^{\pm\mathrm{N/A}}$ 
& - 
& - 
& $1.021^{\pm\mathrm{N/A}}$ 
& $4.098^{\pm\text{N/A}}$ 
& \cellcolor{red!13}$9.513^{\pm\mathrm{N/A}}$ \\

& LMM~\cite{zhang2024large}      
& 90M
& $0.496^{\pm.002}$ 
& $0.685^{\pm.002}$ 
& $0.785^{\pm.002}$ 
& $0.415^{\pm.002}$ 
& $3.087^{\pm.012}$ 
& $9.176^{\pm.074}$ \\

& MCM~\cite{ling2024mcm}    
& -
& \cellcolor{yellow!16}$0.502^{\pm.002}$ 
& $0.692^{\pm.004}$
& $0.788^{\pm.006}$ 
& \cellcolor{red!13}$0.053^{\pm.007}$ 
& $3.037^{\pm.003}$ 
& $9.585^{\pm.082}$  
\\

& MotionCraft~\cite{bian2025motioncraft}    
& 77M
& $0.501^{\pm.003}$ 
& \cellcolor{yellow!16}$0.697^{\pm.003}$ 
& \cellcolor{yellow!16}$0.796^{\pm.002}$ 
& $0.173^{\pm.002}$ 
& \cellcolor{yellow!16}$3.025^{\pm.008}$  
& \cellcolor{yellow!16}$9.543^{\pm.098}$   \\

& GENMO~\cite{li2025genmo}    
& 504M
& - 
& - 
& $0.632^{\pm\mathrm{N/A}}$ 
& $0.216^{\pm\mathrm{N/A}}$ 
& $3.466^{\pm\mathrm{N/A}}$  
& $11.342^{\pm\mathrm{N/A}}$   \\

& Odoriko (Ours)
& 44M
& \cellcolor{red!13}0.512$^{\pm{.002}}$ 
& \cellcolor{red!13}0.709$^{\pm{.003}}$ 
& \cellcolor{red!13}0.805$^{\pm{.002}}$ 
& \cellcolor{yellow!16}0.103$^{\pm{.004}}$ 
& \cellcolor{red!13}2.930$^{\pm{.010}}$ 
& 9.672$^{\pm{.070}}$   \\
\bottomrule
\end{tabular}

    }

    \label{exp:humanml}
\vspace{-20pt}
\end{table*}

\subsection{Comparison on Task-Specific Benchmarks}
\noindent
\textbf{Text-to-Motion.}
We evaluate Odoriko on the HumanML3D benchmark and compare it with recent state-of-the-art methods. 
For fair comparison, we convert our generated motion representation into SMPL parameters and subsequently into the HumanML3D format using the official conversion pipeline.
During evaluation, Odoriko is conditioned on the text prompts, gender, and $\boldsymbol{\beta}$ parameters provided in the HumanML3D test split. 
For samples originating from HumanAct12, which do not include shape annotations, we follow the dataset convention and set the gender to \textit{male} and $\boldsymbol{\beta}$ to a zero vector, corresponding to a canonical body shape.

In Table~\ref{exp:humanml}, Odoriko achieves leading performance among multimodal motion generators (e.g., MotionCraft~\cite{bian2025motioncraft}, GENMO~\cite{li2025genmo}) and remains competitive with recent task-specific text-to-motion specialists \cite{dai2024motionlcm, pinyoanuntapong2024mmm, guo2024momask, uchida2025mola} despite jointly modeling multiple modalities.

Notably, Odoriko outperforms our primary competitor, GENMO, while using substantially fewer parameters, achieving efficiency comparable to single-modality models such as MoMask~\cite{guo2024momask} and MoLA~\cite{uchida2025mola}.
Moreover, Odoriko attains the best results on most metrics while operating under a representation mismatch with the HumanML3D evaluation space—a discrepancy reported to degrade evaluation scores~\cite{li2025genmo}—whereas prior works such as MCM~\cite{ling2024mcm} operate directly in HumanML3D’s native representation.

We attribute this improvement in part to explicit shape conditioning.
Most prior text-to-motion methods condition solely on text and thus absorb inter-subject body-shape variation into the learned motion prior~\cite{zhang2024motiondiffuse, tevet2022human, guo2022generating, dai2024motionlcm, pinyoanuntapong2024mmm, guo2024momask, uchida2025mola}.
Because AMASS comprises motion capture from subjects with diverse body morphologies, shape-dependent kinematics can become entangled with semantic motion patterns in such implicit representations.
By explicitly conditioning on gender and SMPL shape coefficients $\boldsymbol{\beta}$, Odoriko encourages a clearer separation between morphology and semantic intent, allowing the model to learn motion content without compensating for shape-induced variation.

\vspace{4pt}
\noindent
\textbf{Music-to-Dance.}
We evaluate Odoriko on the FineDance benchmark. 
Unlike text-to-motion, music-to-dance generation prioritizes rhythmic precision and dynamic (temporal) coherence over semantic alignment. 
Since FineDance does not provide ground-truth body shape annotations, we replace the shape token with a learnable embedding, without explicitly conditioning on dancer-specific gender or $\boldsymbol{\beta}$ parameters.

As shown in Table~\ref{tab:finedance}, Odoriko achieves the best performance among multimodal baselines in dance quality metrics (FID$_k$ and FID$_g$), and further surpasses recent dedicated music-to-dance methods in BAS. 
These results indicate that our shape-aware architecture does not hinder rhythmic fidelity; instead, it enhances motion realism while maintaining strong music–motion alignment.

\begin{table*}[t]
\centering
\caption{Comparison of motion quality and diversity on the FineDance benchmark. Lower is better for FID; higher is better for BAS and Diversity. Task-specific music-to-dance methods are shown in {\color{gray}{gray}} for reference. \colorbox{red!13}{Best} and \colorbox{yellow!16}{second-best} results are highlighted. $^{*}$Results computed using official checkpoints. 
}
\vspace{-10pt}

\resizebox{0.6\textwidth}{!}{
\begin{tabular}{clcccccccc}
\toprule
\multirow{2}{*}{Category} & \multirow{2}{*}{Method} 
& \multicolumn{2}{c}{Motion Qual.}
&
& \multicolumn{2}{c}{Motion Div.} 
& 
&\multirow{2}{*}{BAS $\uparrow$} \\
\cmidrule{3-4} \cmidrule{6-7}
 &  & FID$_k$ $\downarrow$ 
 & FID$_g$ $\downarrow$ 
 &
 & Div$_k$ $\uparrow$ 
 & Div$_g$ $\uparrow$ 
 &  \\
\midrule

N/A & Real 
& - & - 
&
& 9.73 & 7.44 
&& 0.2120
\\
\midrule

\multirow{6}{*}{\textbf{M2D-only}}
& \color{gray}{FACT \cite{li2021ai}}
& \color{gray}113.38
& \color{gray}97.05
&
& \color{gray}3.36
& \color{gray}6.37
&& \color{gray}0.1831 
\\

& \color{gray}{MNet \cite{kim2022brand}}
& \color{gray}104.71
& \color{gray}90.31
&
& \color{gray}3.12
& \color{gray}6.14
&& \color{gray}0.1864 
\\

& \color{gray}{Bailando \cite{siyao2022bailando}}
& \color{gray}82.81
& \color{gray}28.17
&
& \color{gray}7.74
& \color{gray}6.25
&& \color{gray}0.2029 
\\

& \color{gray}{EDGE \cite{tseng2023edge}}
& \color{gray}94.34
& \color{gray}50.38
&
& \color{gray}8.13
& \color{gray}6.45
&& \color{gray}0.2116 
\\

& \color{gray}{LODGE \cite{li2024lodge}}
& \color{gray}45.56
& \color{gray}34.29
&
& \color{gray}6.75
& \color{gray}5.64
&& \color{gray}0.2397 
\\

& \color{gray}{LODGE++ \cite{li2024lodge++}}
& \color{gray}40.77
& \color{gray}30.79
&
& \color{gray}5.53
& \color{gray}5.01
&& \color{gray}0.2423 
\\
\midrule

\multirow{4}{*}{\textbf{Multimodal}}
& MotionCraft$^{*}$ \cite{bian2025motioncraft}
& 53.61
& \cellcolor{yellow!16}46.14
&
& \cellcolor{yellow!16}10.74
& \cellcolor{red!13}7.72
&& 0.2143 
\\

& VerMo \cite{ling2024versatilemotion}
& 42.89
& -
&
& \cellcolor{red!13}14.61
& -
&& 0.2162 
\\

& M$^3$GPT \cite{luo2024m}
& \cellcolor{yellow!16}42.66
& -
&
& 8.24
& -
&& \cellcolor{yellow!16}0.2261 
\\

& Odoriko (Ours)
& \cellcolor{red!13}37.73
& \cellcolor{red!13}25.36
&
& 7.33
& \cellcolor{yellow!16}6.49
&& \cellcolor{red!13}0.2496 
\\
\bottomrule
\end{tabular}
}

\vspace{-20pt}
\label{tab:finedance}
\end{table*}

\vspace{4pt}
\noindent
\textbf{Video-to-Motion.}
We evaluate Odoriko on EMDB and 3DPW for camera-centric 3D human pose estimation in Table \ref{tab:embd}. 
On EMDB, Odoriko surpasses the multimodal baseline GENMO, as well as achieving competitive performance compared to recent specialist methods~\cite{sun2023trace, shin2024wham, shen2024world, wang2024tram}. 
On 3DPW, our model attains reasonable accuracy, while retaining the flexibility of multimodal conditioning within a unified diffusion framework.

When we exclude the camera-centric root orientation parameter $\theta_{r}^{l}$ and instead directly calculate global orientation using ground-truth camera trajectories, performance drops significantly. 
This suggests that explicitly modeling camera-centric local orientation—rather than relying on global orientation signals—facilitates more stable and accurate pose reconstruction. 

Similarly, removing 2D keypoint inputs leads to marginal degradation on EMDB but substantially larger drops on 3DPW. 
We attribute this discrepancy to dataset characteristics: EMDB primarily contains single-person videos, whereas 3DPW frequently includes multiple subjects. 
In multi-person scenes, 2D keypoints act as spatial cues that help the model disambiguate the target subject, resulting in more pronounced performance gains on 3DPW.

\begin{table*}[t]
\centering
\caption{
Camera-space human motion estimation on EMDB-1 and 3DPW test split.
\colorbox{red!13}{Best} and \colorbox{yellow!16}{second-best} results are highlighted.
}
\vspace{-10pt}

\resizebox{0.9\textwidth}{!}{
\begin{tabular}{clccccccccc}
\toprule
\multirow{2}{*}{Category} 
& \multirow{2}{*}{Method} 
& \multicolumn{4}{c}{EMDB (24)} 
& 
& \multicolumn{4}{c}{3DPW (14)} \\
\cmidrule{3-6} \cmidrule{8-11}
& 
& PA-MPJPE $\downarrow$
& MPJPE $\downarrow$
& PVE $\downarrow$
& Accel $\downarrow$
&
& PA-MPJPE $\downarrow$
& MPJPE $\downarrow$
& PVE $\downarrow$
& Accel $\downarrow$\\
\midrule

\multirow{4}{*}{\textbf{V2M-only}}
& TRACE \cite{sun2023trace}
& 70.9 
& 109.9 
& 127.4 
& 25.5 
&
& 50.9 
& 79.1 
& 95.4 
& 28.6 \\

& WHAM \cite{shin2024wham}
& 50.4 
& 79.7 
& 94.4 
& 5.3 
&& 35.9 
& 57.8 
& 68.7 
& 6.6 \\

& GVHMR \cite{shen2024world}
& 42.7 
& 72.6 
& 84.2 
& \cellcolor{red!13}3.6 
&& 36.2 
& \cellcolor{yellow!16}55.6 
& \cellcolor{yellow!16}67.2 
& 5.0 \\

& TRAM \cite{wang2024tram}
& 45.7 
& 74.4 
& 86.6 
& 4.9 
&& \cellcolor{yellow!16}35.6 
& 59.3 
& 69.6 
& \cellcolor{yellow!16}4.9 \\

\midrule

\multirow{4}{*}{\textbf{Multimodal}}
& GENMO \cite{li2025genmo}
& 42.5 & 73.0 & 84.8 & \cellcolor{yellow!16}3.8 
&& \cellcolor{red!13}34.6 & \cellcolor{red!13}53.9 & \cellcolor{red!13}65.8 & \cellcolor{red!13}4.8 \\

\cmidrule{3-11}

& Ours (w/o $\theta_{r}^{l})$
&  \cellcolor{yellow!16}41.1& 285.9 &345.9  & 7.2
&& 41.1 & 255.8 & 287.4 & 8.6 \\

& Ours (w/o 2D kpts)
& 46.6& 71.1\cellcolor{yellow!16} & \cellcolor{yellow!16}83.7 &9.4 
&&  52.8& 100.2& 114.4 &16.3 \\

& Ours 
& \cellcolor{red!13}41.1 & \cellcolor{red!13}70.2 & \cellcolor{red!13}82.9 & 7.5 
&& 41.1 & 69.9 & 82.0 & 8.3 \\

\bottomrule
\end{tabular}
}

\label{tab:embd}
\vspace{-20pt}
\end{table*}

\subsection{Shape-Aware Motion Generation Evaluation}

We further evaluate whether generated motions faithfully reflect the underlying body shape and gender attributes. 
Unlike conventional motion quality metrics (e.g., FID, diversity), this evaluation explicitly measures the degree to which shape-dependent motion patterns are preserved in the synthesized sequences.

\vspace{4pt}
\noindent
\textbf{Text-to-Motion.}

\vspace{2pt}
\noindent
\emph{Shape Attribute Predictors.}
We train three Transformer-based predictors on the HumanML3D training set and freeze them during evaluation:
a gender classifier that infers biological gender from root-relative joint trajectories;
a continuous $\boldsymbol{\beta}$ regressor via direct regression;
and a discretized $\boldsymbol{\beta}$ classifier that partitions each dimension into 10 bins for categorical prediction.

\vspace{2pt}
\noindent
\emph{Evaluation Metrics.}
Using the above predictors, we define four shape-reflection metrics:
\begin{itemize}
    \item \textbf{Gender Accuracy}: classification accuracy of predicted gender,
    \item \textbf{$\boldsymbol{\beta}$-bin Accuracy}: classification accuracy over discretized shape bins,
    \item \textbf{$\boldsymbol{\beta}$ MAE}: mean absolute error between predicted and ground-truth $\boldsymbol{\beta}$,
    \item \textbf{$\boldsymbol{\beta}$ PVE}: per-vertex error induced by predicted $\boldsymbol{\beta}$.
\end{itemize}
These metrics collectively measure whether the generated motion encodes shape-dependent dynamics consistent with the intended subject.

\vspace{2pt}
\noindent
\emph{Baseline Evaluation Protocol.}
For text-to-motion methods~\cite{tevet2022human, zhang2023generating, guo2024momask, uchida2025mola} trained on HumanML3D, explicit shape conditioning is generally not supported. 
Nevertheless, since HumanML3D retargets all motions to a single canonical body shape, we use that canonical gender and $\boldsymbol{\beta}$ parameters as reference labels for evaluation.
For ShapeMove~\cite{liao2025shape}, which conditions on shape information extracted from text prompts, we follow their prompt construction strategy for fair comparison. 
Specifically, we generate prompts corresponding to the ground-truth gender and $\boldsymbol{\beta}$ of each test sample using SHAPY~\cite{choutas2022accurate}, consistent with their training protocol.

\vspace{2pt}
\noindent
\emph{Results.}
As shown in Table~\ref{tab:shape_eval}, Odoriko significantly outperforms both non-shape-aware and existing shape-aware baselines across all shape-reflection metrics. 
Combined with the motion quality improvements reported in Table~\ref{exp:humanml}, these results demonstrate that Odoriko not only generates high-quality motions but also faithfully preserves subject-specific shape characteristics in the synthesized outputs.

\begin{table*}[t]
\centering
\begin{minipage}[t]{0.54\textwidth}
\centering
\caption{
Shape attribute evaluation on HumanML3D test set.
Classification metrics (Gender Acc.\ and $\boldsymbol{\beta}$-bin Acc.) favor higher values;
regression metrics ($\boldsymbol{\beta}$ PVE and MAE) favor lower.
SA: shape-awareness during generation.
}
\vspace{-10pt}

\resizebox{\textwidth}{!}{%
\begin{tabular}{lccccc}
\toprule
Method 
& SA
& Gender Acc. $\uparrow$
& $\boldsymbol{\beta}$-bin (10) $\uparrow$
& $\boldsymbol{\beta}$ PVE $\downarrow$
& $\boldsymbol{\beta}$ MAE $\downarrow$ \\
\midrule
Real      & -  & 96.9  & 93.0  & 1.88 & 0.102 \\
\midrule
MDM \cite{tevet2022human}       & \color{red}{\xmark} & \cellcolor{yellow!16}80.1 & 41.9 & 8.91 & \cellcolor{yellow!16}0.526 \\
T2M-GPT \cite{zhang2023generating} & \color{red}{\xmark} & 68.2 & 41.5 & \cellcolor{yellow!16}7.71 & 0.566 \\
MoMask \cite{guo2024momask}    & \color{red}{\xmark} & 78.6 & 26.3 & 9.06 & 0.614 \\
MoLA \cite{uchida2025mola}     & \color{red}{\xmark} & 65.6 & 37.8 & 8.26 & 0.592 \\
\midrule
ShapeMove \cite{liao2025shape} & \color{green!50!black}{\cmark} & 74.5 & \cellcolor{yellow!16}52.0 & 8.67 & 0.574 \\
Ours & \color{green!50!black}{\cmark}
& \cellcolor{red!13}92.0
& \cellcolor{red!13}71.8
& \cellcolor{red!13}4.42
& \cellcolor{red!13}0.279 \\
\bottomrule
\end{tabular}%
}

\label{tab:shape_eval}
\end{minipage}
\vspace{-10pt}
\hfill
\begin{minipage}[t]{0.43\textwidth}
\centering
\caption{
Gender preservation on AIST++ test split. 
D: Gender tendency direction, M<F (female > male) or M>F (male > female); 
Pres.: Preservation of biomechanical tendency.
}
\vspace{-10pt}
\setlength{\tabcolsep}{3pt}
\resizebox{\textwidth}{!}{%

\begin{tabular}{lccccccc}
\toprule
\multirow{2}{*}{Metric} &
\multicolumn{3}{c}{Real} &
\multicolumn{3}{c}{Generated} &
\multirow{2}{*}{Pres.} \\
\cmidrule(lr){2-4} \cmidrule(lr){5-7}
& M & F & D & M & F & D & \\
\midrule
HEA     $\rightarrow$& 0.043 & 0.063 & M<F & 0.112 & 0.122 & M<F & \color{green!50!black}{\cmark} \\
HSA     $\rightarrow$& 0.085 & 0.156 & M<F & 0.247 & 0.251 & M<F & \color{green!50!black}{\cmark} \\
MCD     $\rightarrow$ & 0.487 & 0.885 & M<F & 0.545 & 0.709 & M<F & \color{green!50!black}{\cmark} \\
\midrule
Gender Acc. $\uparrow$& \multicolumn{3}{c}{88.5} & \multicolumn{3}{c}{61.5} & - \\
\bottomrule
\end{tabular}%

}

\label{tab:m2d_shape}
\end{minipage}
\vspace{-10pt}
\end{table*}

\vspace{4pt}
\noindent
\noindent
\textbf{Music-to-Dance.}
\noindent
Although music-to-dance generation is primarily driven by tempo and genre, dance motions may still exhibit latent kinematic traits associated with dancer morphology. 
To assess whether such traits are preserved, we evaluate generated motions on the AIST++ test split, which provides paired music, dance, and gender annotations.

We adopt two complementary evaluation strategies.
First, following text-to-motion protocols, we train a Transformer-based gender classifier on joint coordinates from the AIST++ training set and evaluate it on both real and generated dances.
Second, we compute three biomechanics-inspired metrics from joint coordinates—hip sagittal ellipse amplitude (HEA), hip sagittal semi-major axis (HSA), and mediolateral center-of-mass displacement range (MCD)—based on prior biological motion studies~\cite{troje2002decomposing, hsue2014effects}.

As shown in Table~\ref{tab:m2d_shape}, generated dances exhibit gender-consistent patterns in these biomechanical metrics while achieving gender classification accuracy approaching that of real data. 
These results suggest that Odoriko retains morphology-related kinematic tendencies, indicating shape-aware generation even under music-driven conditions.

\vspace{4pt}
\noindent
\textbf{Qualitative results.}
\noindent
In Fig.~\ref{fig:qual}, we show that, under identical text or music conditions, Odoriko generates distinct motions when the input shape parameters change. 
For example, different body morphologies lead to variations in jump height and balance during walking on a narrow plank. 
Similarly, in music-to-dance generation, the model produces stylistic variations under identical audio when conditioned on different shape inputs. 
These results indicate that Odoriko produces motions that consistently reflect the provided bio-morphological attributes.

\subsection{Ablation Study}
\noindent
\textbf{Effectiveness of Shape-Awareness.}
To verify that Odoriko truly leverages subject-specific morphology—rather than generating shape-agnostic yet plausible motions—we perform ablations on both generation and estimation tasks.

\vspace{2pt}
\noindent
\emph{Text-to-Motion.}
We assess the role of shape conditioning by replacing the ground-truth gender and $\boldsymbol{\beta}$ with a canonical body (\textit{male}, $\boldsymbol{\beta}=\mathbf{0}$) while keeping the same text prompts.
As shown in Table~\ref{tab:ablation_shape}, this substitution degrades FID and R-Precision (R3), indicating that removing subject-specific morphology harms both motion realism and text–motion alignment.
This confirms that explicit shape conditioning is essential: Odoriko meaningfully encodes gender and body shape into the synthesized motion.

\vspace{2pt}
\noindent
\emph{Video-to-Motion.}
For motion estimation, we replace the predicted gender and $\boldsymbol{\beta}$ with a neutral gender template and zero $\boldsymbol{\beta}$ at evaluation time.
As reported in Table~\ref{tab:ablation_shape}, this leads to consistent degradation across pose metrics, demonstrating the importance of accurate shape estimation.
In contrast, substituting predicted shape with ground-truth parameters yields only marginal gains over our predictions, suggesting that Odoriko estimates morphology reliably in practice.

\vspace{4pt}
\noindent
\textbf{Inference Steps.}
As shown in Table~\ref{tab:step}, performance improves with more denoising steps and saturates around 25, beyond which no consistent gain is observed.
Among samplers, UniPC~\cite{zhao2023unipc} outperforms DDIM~\cite{songdenoising} at the same step budget.
\begin{table*}[t]
\centering

\begin{minipage}[t]{0.64\textwidth}
\caption{Ablation on shape-awareness across motion estimation (left) and motion generation (right).}
\label{tab:ablation_shape}
\end{minipage}
\hfill
\begin{minipage}[t]{0.32\textwidth}
\caption{FID across samplers and steps.}
\label{tab:step}
\end{minipage}
\vspace{-10pt}

\begin{minipage}[t]{0.44\textwidth}
\centering
\setlength{\tabcolsep}{4pt}
\resizebox{\textwidth}{!}{%
\begin{tabular}{lcccc}
\toprule
\multirow{2}{*}{Method} & \multicolumn{4}{c}{EMDB (24)} \\
\cmidrule(lr){2-5}
 & PA-MPJPE $\downarrow$& MPJPE $\downarrow$& PVE $\downarrow$& Accel $\downarrow$\\
\midrule
GT shape                & \cellcolor{red!13}41.1 & \cellcolor{red!13}70.0 &\cellcolor{red!13} 82.8 &\cellcolor{red!13} 7.5 \\
\midrule
\textit{neutral} gender & 42.2 & 76.3 & 89.3 & 7.6 \\
\textit{zero}-$\boldsymbol{\beta}$ & 41.4 & 72.7 & 86.0 & 7.6 \\
Estimated               & \cellcolor{yellow!16}41.1& \cellcolor{yellow!16}70.2 & \cellcolor{yellow!16}82.9 & \cellcolor{yellow!16}7.5 \\
\bottomrule
\end{tabular}%
}
\end{minipage}
\hfill
\begin{minipage}[t]{0.20\textwidth}
\centering
\setlength{\tabcolsep}{4pt}
\resizebox{\textwidth}{!}{%
\begin{tabular}{lcc}
\toprule
\multirow{2}{*}{Shape} & \multicolumn{2}{c}{HumanML3D} \\
\cmidrule(lr){2-3}
 & R3\,$\uparrow$ & FID\,$\downarrow$ \\
\midrule
GT  &\cellcolor{red!13} 0.805 & \cellcolor{red!13}0.103 \\
\midrule
Cano. & \cellcolor{yellow!16}0.768 & \cellcolor{yellow!16}0.592 \\
\bottomrule
\end{tabular}%
}
\end{minipage}
\hfill
\begin{minipage}[t]{0.32\textwidth}
\centering
\setlength{\tabcolsep}{4pt}
\resizebox{\textwidth}{!}{%
\begin{tabular}{llcc}
\toprule
% Sampler & Step & T2M\,$\downarrow$ & M2D\,$\downarrow$ \\
\multirow{2}{*}{Sampler} & \multirow{2}{*}{Step} & HumanML3D & FineDance \\
\cmidrule{3-4}
& & FID\,$\downarrow$  & FID$_{k}$\,$\downarrow$ \\
\midrule
\multirow{5}{*}{UniPC} 
      & 1  & 1.804 & 303.42\\
      & 5  & 0.906& 38.65\\
      & 10 & 0.200& 37.96\\
      & 25 & \cellcolor{red!13}0.103& \cellcolor{red!13}37.73\\
      & 50 & 0.113& 37.85\\
\midrule
DDIM  & 25 & \cellcolor{yellow!16}0.107& \cellcolor{yellow!16}37.74\\
\bottomrule
\end{tabular}%

}
\end{minipage}

\vspace{-20pt}
\end{table*}

\vspace{-4pt}
\section{Conclusion}
In this work, we introduced Odoriko, a unified shape-aware multimodal diffusion framework for human motion modeling.
Unlike prior approaches that assume a canonical body shape, Odoriko explicitly incorporates biological gender and SMPL shape parameters $\boldsymbol{\beta}$ as conditioning signals within a multimodal architecture. We proposed a Shape-Aware Motion Transformer that integrates shape tokens with heterogeneous modalities (text, music, video, and 2D keypoints) via hybrid token-wise and global conditioning, enabling joint reasoning over motion dynamics, multimodal context, and subject-specific morphology.
Odoriko further unifies motion generation and camera-centric human pose estimation within a single diffusion backbone. Extensive experiments demonstrate competitive performance across tasks, with consistent improvements in shape consistency and controllability.

% \clearpage\mbox{}Page \thepage\ of the manuscript.
% \clearpage\mbox{}Page \thepage\ of the manuscript.
% \clearpage\mbox{}Page \thepage\ of the manuscript.
% \clearpage\mbox{}Page \thepage\ of the manuscript.
% \clearpage\mbox{}Page \thepage\ of the manuscript. This is the last page.
% Now we have reached the maximum length of an ECCV \ECCVyear{} submission (excluding references and acknowledgements).
% References should start immediately after the main text, but can continue past p.\ 14 if needed. 
% \clearpage  % TODO FINAL: This \clearpage needs to be removed from both review and camera-ready versions.

% \section*{Acknowledgements}
% Please insert your acknowledgments here.

\bibliographystyle{splncs04}
\bibliography{main}

\appendix

\section{Implementation Details}
\subsection{Network Details}

\begin{table}[h]
    \centering
    \caption{Transformer architecture details of Odoriko.}
    \begin{tabular}{c|c}
    \toprule
        Latent dim & 512 \\
        Feedforward (FF) dim & 1024 \\
        CLIP feature dim & 512 \\
        T5 feature dim & 768 \\
        EDGE feature dim & 512 \\
        TRAM feature dim & 1024 \\
        \# attention heads & 4 \\
         \# multimodal blocks & 8 \\
        \# motion refinement blocks & 8 \\
    \bottomrule
    \end{tabular}
    \label{tab:detail}
\end{table}

Odoriko adopts a Transformer~\cite{vaswani2017attention}-based architecture as the backbone of the diffusion model. 
All modality-specific features are projected into a shared latent space before being processed by the transformer blocks.

For the motion input $\mathbf{x}_t$, each frame is represented by a 136-dimensional vector, i.e., $\mathbf{x}_t \in \mathbb{R}^{136}$. 
The first four dimensions encode the root joint dynamics, while the remaining 132 dimensions represent the rotations of the simplified SMPL \cite{loper2015smpl} skeleton with 22 joints used in the HumanML3D \cite{guo2022generating} representation. 
Specifically, we use the 6D rotation \cite{zhou2019continuity} for each joint, resulting in $4 + (22-1)\times6 = 136$ dimensions in total.

Each modality encoder produces features with different dimensionalities. 
Specifically, we use CLIP~\cite{radford2021learning} and T5~\cite{raffel2020exploring} for text encoding, EDGE~\cite{tseng2023edge} for music encoding, and TRAM~\cite{wang2024tram} for video feature extraction. 
To unify these heterogeneous representations, each modality feature is passed through a linear projection layer (adapter) that maps the feature into the shared latent space of dimension 512. 
The overall architectural specifications are summarized in Table~\ref{tab:detail}.

For the video-to-motion estimation task, the motion representation slightly differs from the generation tasks because it includes camera-centric root rotation $\theta_r^{l} \in \mathbb{R}^{6}$. 
To handle this representation, we introduce an additional lightweight transformer module dedicated to this task. 
This module consists only of motion refinement blocks, as no text tokens are required. 
To keep the computational cost small, the module uses a reduced latent dimension of 256 and 4 transformer layers.

Following MDM~\cite{tevet2022human}, we employ absolute positional encoding (PE) and concatenate the diffusion timestep embedding to the input tokens. 
This design still remains commonly used in motion diffusion models and has been shown to provide stable training for motion sequence generation \cite{tevet2022human, lee2025move, liao2025shape}.

\subsection{Training Details}

We train Odoriko using the AdamW optimizer \cite{loshchilov2019decoupled} with a fixed learning rate of $1\times10^{-4}$ and momentum parameters $\beta_1=0.9$ and $\beta_2=0.9999$. 
An exponential moving average (EMA) of the model parameters is maintained during training with a decay rate of 0.9999.

Thanks to the efficient architecture of Odoriko compared to recent  DiT \cite{esser2024scaling}-based models such as GENMO \cite{li2025genmo}, the model can be trained on a single NVIDIA A6000 GPU. 
In our experiments, we use a batch size of 128 and train the model for approximately 8 days.

During training, we adopt different maximum motion sequence lengths depending on the task. 
For text-to-motion, the maximum sequence length is set to 196 following MDM. 
For music-to-dance, we use a maximum length of 150 following EDGE. 
For video-to-motion estimation, which focuses on accurate frame-wise pose reconstruction, we use a shorter sequence length of 16 frames.

To handle the discrepancy in motion lengths across tasks within a unified framework, all sequences are padded to a fixed length of 196. 
Specifically, sequences shorter than this length are zero-padded, and the padded regions are ignored using masked self-attention so that they do not affect the training dynamics.

For the diffusion process, we follow the standard guided diffusion training paradigm~\cite{dhariwal2021diffusion}. 
We adopt a cosine noise schedule and use 50 denoising steps, following the training configuration of MDM, which employs a relatively small number of diffusion steps for motion generation in the raw motion domain.

\subsection{Classifier-Free Guidance}

To enable Classifier-Free Guidance (CFG) \cite{ho2021classifier} during inference, we adopt the standard training strategy of randomly dropping conditioning information. 
Specifically, during training for generation tasks (text-to-motion and music-to-dance), the multimodal conditioning features $\mathbf{c}$ are replaced with a learnable null embedding $\varnothing$ with a probability of 0.1. 
This allows the model to learn both conditional and unconditional predictions, which are required for CFG at inference time.

In contrast, for the reconstruction task (video-to-motion), the conditioning inputs—video features and 2D keypoints—are always preserved during training. 
Since the goal of this task is to reconstruct motion consistent with the given visual observations, unconditional sampling is unnecessary, and CFG-based inference is therefore not applied.

During inference, the CFG prediction is computed as:
\begin{equation}
\hat{\mathbf{x}}_0(x_t, t, \mathbf{c}, \mathbf{s})
=
f_\theta(x_t, t, \varnothing, \mathbf{s})
+
w \big(f_\theta(x_t, t, \mathbf{c}, \mathbf{s}) - f_\theta(x_t, t, \varnothing, \mathbf{s})\big).
\end{equation}
where $f_\theta$ denotes the Odoriko model, $\mathbf{c}$ represents the multimodal conditioning features, and $\mathbf{s}$ denotes the shape-related conditioning signals. 
The scalar $w$ controls the strength of the guidance. 
Following MDM, which applies CFG in the data-domain motion generation setting, we set $w=2.5$ for generation tasks (text-to-motion and music-to-dance). 
For the video-to-motion task, we set $w=1$, which effectively disables guidance and corresponds to standard conditional diffusion inference.

\section{Standard Benchmark Evaluation Protocol Details}

\subsection{Text-to-Motion}

Following prior works~\cite{tevet2022human, guo2024momask, uchida2025mola}, we evaluate on the HumanML3D test split and repeat the evaluation 20 times, reporting the mean and standard deviation for all metrics.

We adopt standard text-to-motion evaluation metrics.
Frechet Inception Distance (FID) measures motion realism by comparing the feature distributions of generated and real motions.
R-Precision evaluates text–motion alignment by measuring retrieval accuracy between motion and text embeddings.
Multimodal Distance (MM-Dist) quantifies the average embedding distance between generated motions and their corresponding text descriptions.
Finally, Diversity measures the variability among motions generated from different text prompts.

\subsection{Music-to-Dance}

Following LODGE~\cite{li2024lodge}, we evaluate the quality of generated dances on FineDance \cite{li2023finedance} test split using both kinematic and geometric motion features.
Specifically, we compute FID and Diversity (Div) in two feature spaces:
(i) kinematic features, denoted as FID$_k$ and Div$_k$, which capture physical motion dynamics, and
(ii) geometric features, denoted as FID$_g$ and Div$_g$, which reflect overall dance choreography.
Both feature representations are derived from SMPL joint coordinates.

To evaluate rhythmic consistency between generated dance and the input music, we compute the Beat Alignment Score (BAS)~\cite{li2021ai, li2024lodge}, which measures how well motion beats align with musical beats.

\subsection{Video-to-Motion}

Video-to-motion (local 3D pose estimation) benchmarks such as EMDB \cite{kaufmann2023emdb} and 3DPW \cite{von2018recovering} evaluate motion using the 24-joint SMPL skeleton, where joint coordinates are represented in a root-relative coordinate system by subtracting the global root translation, thereby focusing the evaluation on local pose estimation.

Odoriko generates motion sequences represented with 22 joints following the HumanML3D convention.
For evaluation, we reconstruct the corresponding SMPL representation using the gender and SMPL shape parameters $\boldsymbol{\beta}$ predicted by Odoriko and compute the SMPL joint coordinates.

To comply with the 24-joint evaluation protocol, the two additional joints (left and right hands), which are not included in the 22-joint representation, are zero-padded during evaluation. As these joints have minimal influence on overall motion dynamics, their absence has negligible impact on the evaluation results.

\section{Shape-Awareness Evaluation Protocol Details}

\subsection{Text-to-Motion}

This section provides additional details on the evaluation protocol used to measure shape awareness in the text-to-motion task, complementing Sec.~\textcolor{red}{4.3} of the main manuscript.

\subsubsection{Evaluator Design.}
To assess whether generated motions reflect subject morphology, we train three Transformer-based evaluators: (1) a gender classifier, (2) a $\boldsymbol{\beta}$ bin classifier, and (3) a $\boldsymbol{\beta}$ regressor. 
All evaluators share the same architecture, consisting of an 8-layer Transformer with 4 attention heads and a latent dimension of 512.

The input to each evaluator is a motion sequence represented by 3D joint coordinates with a maximum length of 196 frames. 
Sequences shorter than this length are zero-padded, and padded tokens are masked during self-attention. 
The models differ only in the output dimension of the prediction head attached to the classification token. 
The gender classifier predicts a 2-dimensional output, while the $\boldsymbol{\beta}$ bin classifier and the $\boldsymbol{\beta}$ regressor produce 10-dimensional outputs corresponding to the bin indices and the SMPL shape parameters, respectively.

\subsubsection{$\boldsymbol{\beta}$ Bin Construction.}
For the $\boldsymbol{\beta}$ bin classifier, we first collect all $\boldsymbol{\beta}$ vectors from the HumanML3D training split and reduce their dimensionality from 10 to 3 using principal component analysis (PCA). 
K-means clustering is then performed in this reduced space to obtain 10 clusters using Euclidean distance. 
Each $\boldsymbol{\beta}$ vector is assigned to its nearest cluster, producing 10 discrete shape categories used as classification targets.

\subsubsection{Representation for Evaluation.}
Evaluation metrics for motion generation are known to be sensitive to the choice of motion representation space~\cite{guo2022generating, meng2025rethinking}. 
To avoid biases introduced by dataset-specific representations, we perform evaluation directly on 3D joint coordinates rather than representations tied to specific benchmarks such as HumanML3D, MARDM, or SMPL parameter spaces.

Because raw joint coordinates can exhibit large variance due to global translation, we represent motions using root-relative coordinates. 
Specifically, each joint position is expressed relative to the root joint, which stabilizes the input distribution and improves evaluator robustness. The inputs are further normalized using the mean and standard deviation computed from the training set.

\subsection{Music-to-Dance}

For the music-to-dance task, we adopt a similar evaluation protocol. 
A Transformer-based gender classifier is trained to measure whether generated dances preserve gender-related motion characteristics. 
The input representation follows the same root-relative 3D joint coordinate format used in the text-to-motion evaluation.

In addition to classification accuracy, we evaluate morphology-related motion characteristics using three kinematic metrics derived from biological motion studies~\cite{troje2002decomposing, barclay1978temporal}. 
Given a motion sequence represented as joint positions over time $\mathbf{J} \in \mathbb{R}^{T \times 22 \times 3}$ in the SMPL coordinate frame (X: mediolateral, Y: vertical, Z: anteroposterior), we compute the following measures.

\paragraph{Hip Sagittal Ellipse Amplitude (HEA).}
Barclay et al.~\cite{barclay1978temporal} identified hip motion as a primary cue for gender perception in biological motion.
Troje~\cite{troje2002decomposing} further analyzed this phenomenon and showed that both the shoulders and hips exhibit elliptical motions in the sagittal plane, where the amplitudes of these ellipses depend on the widths of the shoulders and pelvis.
We compute the hip center
\[
\mathbf{h}_t = \frac{1}{2}(\mathbf{J}_{t,\text{left\_hip}} + \mathbf{J}_{t,\text{right\_hip}})
\]
and project it onto the sagittal plane
\[
\mathbf{h}^{YZ}_t = [\mathbf{h}_{t,y}, \mathbf{h}_{t,z}].
\]
After mean-centering the trajectory $\tilde{\mathbf{h}}^{YZ}_t = \mathbf{h}^{YZ}_t - \bar{\mathbf{h}}^{YZ}$, we compute the radii $r_t = \|\tilde{\mathbf{h}}^{YZ}_t\|$ and define amplitude as their standard deviation:
\[
\text{HEA} = \sqrt{\frac{1}{T-1}\sum_{t=1}^{T}(r_t - \bar{r})^2}
\]
where $\bar{r} = \frac{1}{T}\sum_{t=1}^{T}r_t$.
Female walkers exhibit larger values due to wider pelvic structure~\cite{troje2002decomposing}.

\paragraph{Hip Sagittal Semi-Major Axis (HSA).}
To further characterize the elliptical hip trajectory, we compute the covariance matrix of the centered sagittal trajectory
\[
\mathbf{C} = \frac{1}{T-1}\sum_{t=1}^{T} \tilde{\mathbf{h}}^{YZ}_t(\tilde{\mathbf{h}}^{YZ}_t)^\top .
\]
Eigenvalue decomposition $\mathbf{C} = \mathbf{V}\mathbf{\Lambda}\mathbf{V}^\top$ yields eigenvalues $\lambda_1 \geq \lambda_2$. 
The semi-major axis length is defined as
\[
\text{HSA} = \sqrt{\lambda_1}.
\]
Prior studies report larger values for female subjects.

\paragraph{Mediolateral Center-of-Mass Displacement Range (MCD).}
Hsue and Su~\cite{hsue2014effects} reported that women exhibit larger mediolateral center-of-mass displacement during locomotion. 
We assume and compute the center of mass as;
\[
\mathbf{c}_t = \frac{1}{22}\sum_{j=1}^{22} \mathbf{J}_{t,j},
\]
extract the mediolateral component $c_t^X = \mathbf{c}_{t,x}$, remove global drift via least-squares linear detrending to obtain $\tilde{c}_t^X$, and measure the peak-to-peak displacement
\[
\text{MCD} = \max_t(\tilde{c}_t^X) - \min_t(\tilde{c}_t^X).
\]

\section{Additional Ablation Study}
\begin{table}[h]
    \centering
        \caption{Ablation study on model architecture.}
    \begin{tabular}{c|c c | c c}
            \toprule
        \multirow{3}{*}{Method} & \multicolumn{2}{c}{Shape Cond.}& \multicolumn{2}{c}{Shape Est.} \\
        \cmidrule{2-5}
                                & \multicolumn{2}{c}{HumanML3D}& \multicolumn{2}{c}{EMDB (24)} \\
         & R3 $\uparrow$ & FID $\downarrow$& PA-MPJPE $\downarrow$ & MPJPE $\downarrow$  \\
         \midrule
       w/o CLIP & \cellcolor{yellow!16}0.799& \cellcolor{red!13}0.100&-&-\\
       $v$ prediction & 0.720&0.335&43.7&76.9 \\
       Non causal& 0.774&  0.118&\cellcolor{yellow!16}41.6&\cellcolor{yellow!16}71.0 \\
       \midrule
       Ours & \cellcolor{red!13}0.805& \cellcolor{yellow!16}0.103&\cellcolor{red!13}41.1&\cellcolor{red!13}70.2\\
       \bottomrule
    \end{tabular}

    \label{tab:abl_app}
\end{table}
First, we examine the effect of incorporating CLIP~\cite{radford2021learning} as a global feature extractor. In this experiment, we remove the CLIP encoder and replace the global feature with the mean-pooled representation of the T5 word-level embeddings. As shown in Table~\ref{tab:abl_app}, removing CLIP results in a noticeable degradation in R-Precision@3 (R3), while slightly improving FID. This trend is consistent with observations reported in prior works~\cite{chang2025casim, maldonado2025moclip}, which show that incorporating CLIP features generally improves text–motion alignment metrics such as R-Precision, sometimes at the cost of a modest degradation in FID. Since the text encoder does not directly influence the video-to-motion task, we do not report video-to-motion metrics for this ablation.

Next, we investigate the impact of the training objective. Recent generative models across multiple modalities—including text-to-image~\cite{esser2024scaling, labs2025flux1kontextflowmatching}, text-to-video~\cite{wan2025wan}, and image-to-audio~\cite{cheng2025mmaudio}—have increasingly adopted Flow Matching (FM)~\cite{lipman2022flow} instead of the standard diffusion objective~\cite{ho2020denoising}. Following this trend, we train a variant of our model using velocity prediction ($v$-prediction) and perform inference with an Euler sampler. 
However, our experiments show that direct $\mathbf{x}_0$ prediction achieves better performance for both motion generation (text-to-motion) and local 3D pose estimation (video-to-motion) in the raw motion domain. 
We hypothesize that this behavior arises because human motion data is highly structured and physically constrained compared to image or video data, making direct $\mathbf{x}_0$ prediction more stable for trajectory generation. Moreover, many recent FM-based models operate in learned latent spaces (e.g., VAE latents), whereas our model generates motions directly in the raw motion space, which may further limit the effectiveness of FM-style objectives.

Finally, we study the design of shape conditioning. In the proposed architecture, shape information is incorporated in a causal manner, where biological gender first conditions the prediction of the SMPL shape parameters $\boldsymbol{\beta}$. To validate this design choice, we train an alternative model that removes this causal structure. Specifically, we concatenate the gender embedding and the $\boldsymbol{\beta}$ embedding into a single vector, project it into a unified shape token, and feed this token into both the multimodal fusion stage and the motion refinement stage. As shown in Table~\ref{tab:abl_app}, this non-causal conditioning strategy degrades performance across tasks, yielding lower R-Precision@3 and higher FID for text-to-motion, as well as higher PA-MPJPE and MPJPE for video-to-motion. 
These results indicate that the proposed causal formulation provides a more effective mechanism for incorporating shape information into the model.

\section{Additional Qualitative Results}

To present additional qualitative results in video form, we provide HTML files that allow reviewers to visualize the results locally.
The webpage structure follows the widely used Nerfies-style project page format \cite{nerfies2021website}.

Although the HTML files may reference a small number of external resources (e.g., fonts or style assets), they are used solely for rendering purposes and do not contain any links to external content related to this work. Therefore, they do not compromise anonymity or allow modifications after the submission deadline.

Please refer to the provided HTML files for further qualitative results. 
All files are included in the same ZIP archive as this supplementary material.

\section{Further Discussion}
\subsection{Discussion on FID in Dual-Mode Training}

In the main manuscript (Sec.~\textcolor{red}{4.2}), we note that FID of Odoriko on the HumanML3D benchmark~\cite{guo2022generating} is slightly higher than that of some recent state-of-the-art text-to-motion methods, which is expected. 
One factor contributing to this gap is the representation mismatch between the motion representation used by our model and the evaluation space of HumanML3D, as also noted in GENMO~\cite{li2025genmo}.
Generated motions must be converted from our internal representation to SMPL and then further transformed into the HumanML3D representation before evaluation.
This additional conversion step inevitably introduces small numerical discrepancies, which can negatively affect metrics such as FID that are sensitive to distributional differences.

Beyond representation mismatch, we believe another contributing factor is the dual-mode training paradigm adopted in our framework.
Odoriko is designed to jointly support two fundamentally different tasks: 
(i) motion generation (e.g., text-to-motion and music-to-dance), and 
(ii) motion estimation (video-to-motion or 3D human pose estimation).

These tasks impose inherently different learning objectives.
Motion generation aims to model a target distribution of plausible motions conditioned on an input signal.
As a result, stochasticity is desirable because multiple valid motions may correspond to the same condition.
In contrast, motion estimation is a regression problem whose objective is to recover a specific ground-truth motion sequence from observations.
In this setting, deterministic predictions that closely match the ground-truth labels are generally preferred.

Unifying these two modes within a single architecture therefore introduces a non-trivial optimization challenge.
The model must simultaneously maintain stochastic generative capability while also producing accurate deterministic predictions for estimation tasks.
Balancing these competing requirements can lead to compromises in optimization, potentially affecting performance on either generation or estimation benchmarks.

Interestingly, a similar phenomenon can be observed in GENMO, which also attempts to unify motion generation (text-to-motion and music-to-dance) and motion estimation (video-to-motion) within a single framework.
GENMO reports strong performance on the estimation side but comparatively weaker performance on certain generation benchmarks.
This trend suggests that the trade-off may arise from the inherent difficulty of jointly optimizing both paradigms within a unified model rather than from a specific architectural design.

Despite these challenges, Odoriko demonstrates competitive performance across both domains while additionally introducing explicit modeling of subject morphology.
As shown in our experiments, Odoriko achieves improved performance compared to GENMO on text-to-motion generation and also demonstrates competitive results on pose estimation benchmarks such as EMDB~\cite{kaufmann2023emdb}.
These results highlight that explicit morphology modeling can be integrated into a unified multimodal framework while maintaining strong performance across heterogeneous tasks.

\subsection{Ethical Statement}

We acknowledge that conditioning motion generation on attributes such as body shape or gender may raise concerns about reinforcing stereotypical patterns in generated motions. 
As with many learning-based approaches, our model reflects statistical patterns present in the training data and may therefore inherit biases contained in the underlying datasets.

To mitigate potential risks, we train Odoriko exclusively on publicly available datasets \cite{guo2022generating, li2021ai, li2023finedance, kaufmann2023emdb, ionescu2013human3, von2018recovering} that are widely used in the research community. 
The generated motions are learned from these curated datasets rather than from uncontrolled or unverified data sources. 
Our intention in incorporating shape-related attributes is to capture biomechanical variations in human motion, such as differences in body morphology, rather than to encode cultural or social stereotypes. 
We also note that the binary gender parameterization used in our framework is an architectural constraint inherited from the SMPL body model and does not represent the full spectrum of gender identity.

Nevertheless, datasets may still contain biases, and we encourage future work to further investigate fairness and bias in motion generation systems, including the development of more diverse datasets and evaluation protocols.

\subsection{Limitation}

Odoriko is, to our knowledge, the first framework that enables multimodal motion generation while explicitly incorporating shape awareness into the motion modeling process. 
However, the shape factor $\mathbf{s}$ used in our framework is derived from the SMPL body model. 
As a result, the range of shapes that can be represented is limited to the parameter space defined by SMPL.
Consequently, the current formulation may not generalize well to characters whose shapes cannot be well approximated by the SMPL model, such as highly stylized or non-human characters (e.g., slime-like creatures, vampires, or dwarfs). 
In such cases, directly conditioning the model on the true body shape becomes challenging.

To address this limitation, a promising future direction is to incorporate shape representations derived directly from 3D mesh geometry, which would allow the model to support a wider range of body structures and enable more general shape-conditioned motion generation.
\end{document}